%% file: main.tex
\documentclass[10pt,letterpaper]{article}

\usepackage{cogsci}
\usepackage{xcolor,soul}

\cogscifinalcopy 

\usepackage{pslatex}
\usepackage[natbibapa]{apacite}
\usepackage{float} 
                   
\usepackage{graphicx}
\graphicspath{{./figs/}}
\usepackage{booktabs}
\usepackage{float}
\usepackage{multirow}
\usepackage{amsmath}
\usepackage{hyperref}
\DeclareMathOperator*{\argmax}{arg\,max}

\title{End-to-end Deep Prototype and Exemplar Models for Predicting Human Behavior}
  
\author{
   Pulkit Singh$^1$, Joshua C. Peterson$^1$, Ruairidh M. Battleday$^1$, Thomas L. Griffiths$^{1,2}$ \\
   $^1$Department of Computer Science, Princeton University\\$^2$Department of Psychology, Princeton University\\
   \texttt{\{pulkit,joshuacp,battleday,tomg\}@princeton.edu}}

\begin{document}
\maketitle

\begin{abstract}
Traditional models of category learning in psychology focus on representation at the category level as opposed to the stimulus level, even though the two are likely to interact. The stimulus representations employed in such models are either hand-designed by the experimenter, inferred circuitously from human judgments, or borrowed from pretrained deep neural networks that are themselves competing models of category learning. In this work, we extend classic prototype and exemplar models to learn both stimulus and category representations jointly from raw input. This new class of models can be parameterized by deep neural networks (DNN) and trained end-to-end. Following their namesakes, we refer to them as Deep Prototype Models, Deep Exemplar Models, and Deep Gaussian Mixture Models. Compared to typical DNNs, we find that their cognitively inspired counterparts both provide better intrinsic fit to human behavior and improve ground-truth classification.

\end{abstract}

\section{Introduction}
Categorization is central to cognition, and so models of category learning are ubiquitous in cognitive science \citep{reed72,posnerk68,nosofsky1984choice,kruschke92}. Most models of category learning can be understood as methods for inferring the structure of different categories (\textit{i.e.,} their representations). For example, prototype models represent a category as an idealization or abstraction---typically the mean---of its members \citep{reed72}, whereas exemplar models represent a category by all previously encountered members \citep{nosofsky1984choice}. These category representations are by definition a function of the underlying stimulus representation, which is typically fixed in advance.

How the modeler should choose the ``right'' stimulus representation for any particular task is not clear. When the representations are designed by hand, they may be considerably biased. A popular alternative is to infer mental representations directly from judgments via multi-dimensional scaling \cite[MDS;][]{shepard80}, but this process is sensitive to context \citep{medingg93}, and when applied to downstream modeling, is limited to predicting behavior \textit{from} behavior.
Further, neither of these options provide a stimulus-to-representation mapping, nor a theory of how humans might acquire one. More problematic, however, is the observation that stimulus representations are unlikely fixed at all, and their structure may be influenced by the process of category learning itself \citep{schynsgt98}.

More recent work \cite[\textit{e.g.,}][]{sanders2018using,peterson2018evaluating,battleday2019capturing,lake2015deep} leverages representations learned from deep neural networks to support psychological models. Unlike hand design or MDS, these networks both model learning and provide a mapping from raw inputs to low-, mid-, and high-level representations \citep{lecun2015deep}. However, these networks are typically trained for the task of categorization in a way that is not motivated by psychological theories. In particular, they are trained to discriminate categories via linear boundaries over their learned feature spaces, and do not explicitly represent the categories themselves. When we use them in service of psychological modeling, we are borrowing from an (implicit) theory of category learning that we do not espouse in order to support a different theory that we do. 

Theories of feature learning in psychology do exist \citep[\textit{e.g.,}][]{rumelhart1985feature,austerweilg13}, but they do not model the interaction between feature learning and category learning, and are often not computationally tractable for large, high-dimensional stimulus sets. While a variety of burgeoning unsupervised deep neural networks \citep[\textit{e.g.,}][]{dumoulin2016adversarially,ji2019invariant,donahue2016adversarial} provide competitive alternate candidates for stimulus representation that could support effective categorization, they are likewise fixed after training. 

To address these shortcomings, we propose a probabilistic framework for incorporating feature learning into classic accounts of category learning, including prototype models, exemplar models, and a family of mixture models that define a continuum between these two extremes. Our framework can be instantiated via deep neural networks, backpropagating through each categorization model (\textit{i.e.,} category representation learner) to a feature learning network that is learned simultaneously. We call these networks Deep Prototype Models (DPM), Deep Exemplar Models (DEM), and Deep Gaussian Mixture Models (DGMM). When trained on a popular benchmark dataset from machine learning, we find that our new family of models provide excellent out-of-sample fit to human uncertainty behavior for a large dataset of over $500,000$ judgments, and boost ground-truth classification accuracy compared to traditional deep neural networks.

\section{Formalization of Category Learning}
\label{sec:formalization}
The problem of category learning can be formalized as inferring the probability that a stimulus, $x$, belongs to each of a set of $C$ categories, $p(c_i\,\in C|\,x)$ \citep{ashbyar95,anderson91}. Bayes' rule implies that
\begin{equation}
    p(c_i\,|\,x) = \frac{p(x\,|\,c_i)p(c_i)}{\sum_j p(x\,|\,c_j)p(c_j)}.
    \label{bayes}
\end{equation}
The prior over categories, $p(c_j)$, can reflect the frequency or saliency of each category. When it is uniform, the posterior is proportional to the likelihood of the stimulus given the category:
\begin{equation}
    p(c_i\,|\,x) \propto p(x\,|\,c_i).
\end{equation}
Our choice of a functional form for the likelihood, $p(x\,|\,c_i)$, specifies the strategy for representing categories, with each of the classical categorization models listed above corresponding to a particular functional form \citep{ashbyar95}. 


A simple first choice for $p(x\,|\,c_i)$ is a Gaussian density, such that stimuli in each category are normally distributed with mean $\mu_i$ and covariance $\Sigma_i$. This implies that
\begin{equation}
    p(c_i\,|\,x) \propto e^{-d(x,\,\,\mu_i,\,\,\Sigma_i)^2},
\end{equation}
%
%
where $d$ is the Mahalanobis distance between the stimulus and the mean of the category given its covariance structure $\Sigma_i$. Gaussian likelihoods correspond to the probabilistic formulation of prototype models over integral psychological spaces, with category prototypes given by ${\bf \mu}_i$, and the similarity of a stimulus to the prototype inversely proportional to its distance. When $\Sigma_i \neq {\bf I}$, new variants of more complex prototype models emerge, including cases that are equivalent to models with curved decision boundaries \citep{ashbym93, minda2001prototypes, battleday2019capturing}. 
%

A much more flexible form for the likelihood is the kernel density estimator, which allows both the distribution of stimuli in each category to be arbitrarily distributed and the boundary between categories to be arbitrarily shaped. In this model, the probability distribution representing a category is composed of the normalized sum of a set of kernel distributions over the stimulus space. Taking a Gaussian kernel, we have that
%
\begin{equation}
    p(c_i\,|\,x) \propto \sum_{x^\prime} e^{-d(x,\,\,x^\prime\!\!, \,\,{\bf I})^2},
\end{equation}
where $x^\prime$ are kernel positions for $c_i$, and $\Sigma_i$ is typically taken to be the identity matrix, ${\bf I}$; this reduces the Mahalanobis distance to the Euclidean distance. When all previously encountered members of category $i$ are used as kernel locations, the Gaussian kernel density estimator is equivalent to the exemplar model \citep{ashbyar95}. Intuitively, this means that the probability that stimulus $x$ comes from category $c_i$ is inversely proportional to the sum of its distances to each member of that category. Other variations of exemplar models may include alternate distance metrics, scaling parameters for the distance metric, and convex attentional transformations of $x$ \citep{nosofsky1984choice}.

Finally, we can consider a class of models that interpolate between these two extremes---of a single Gaussian likelihood for each category and one Gaussian kernel for each exemplar---by allowing the number of kernels per category, $K$, to vary between $1$ and the number of exemplars in that category. This corresponds to a mixture of Gaussians, and allows the modeler to trade off between model accuracy and complexity given data \citep{griffithsscn07}. Although less commonly used, Gaussian mixture models have also been proposed as models of human category learning \citep{rosseel02}, and are special cases of other influential categorization frameworks \citep{griffithscsn07, anderson91}.


\section{Incorporating Feature Learning}
Incorporating feature learning into classic models implies that, in addition to a learnable category representation, a joint model must also specify a learnable transformation of stimulus inputs, $\phi(\cdot)$, over which the categorization model will operate. The idea of inferring representations of stimuli for input into categorization models has been employed for some time via strategies such as MDS \citep{shepard80}, or by searching over a range of feature extraction algorithms \citep{battleday2019capturing}. 
However, although these methods yield a transformed input representation, $\phi(x)$, they do not provide the corresponding transformation $\phi(\cdot)$ itself, nor a means to learn it.
The transformation itself is important because it both explains part of human learning and allows a model to generalize to new stimuli.

\begin{figure*}[!ht]
  \centering
  \includegraphics[width=0.6\linewidth,trim={0mm 0mm 0mm 0mm},clip]{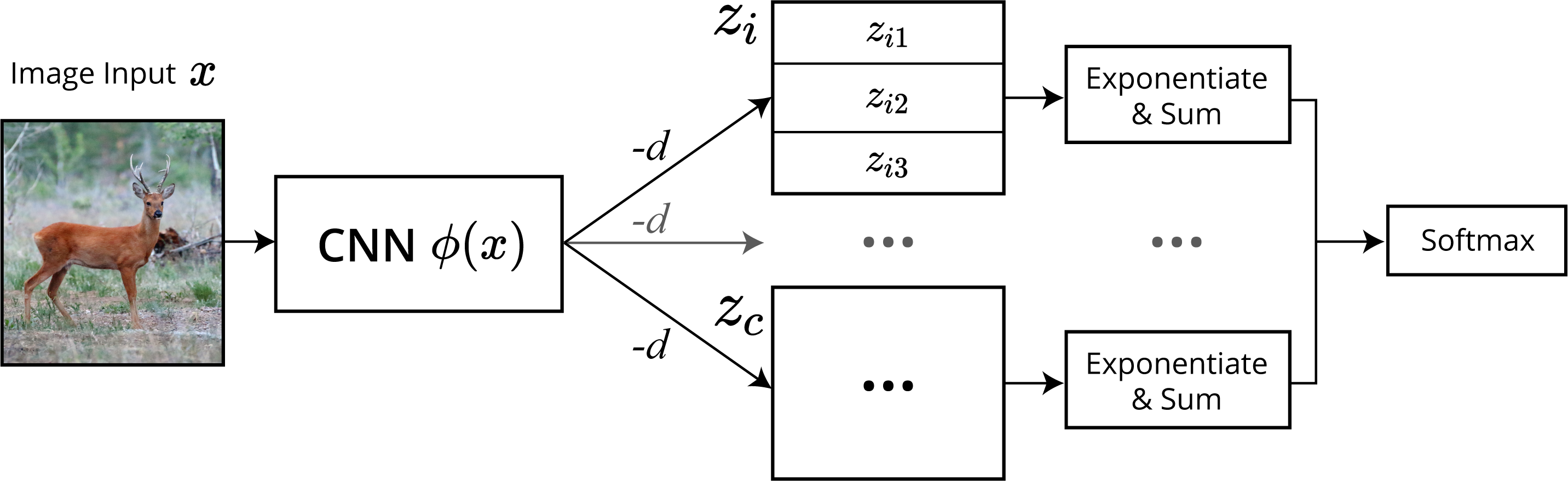}
  \caption{General schematic for deep categorization models.}
  \label{fig:diagram}
\end{figure*}

For complex naturalistic stimuli, we want the flexibility to learn a transformation $\phi(\cdot)$---and thus, a resulting representation $\phi(x)$---that can best support a particular categorization strategy or category representation. However, if $\phi(\cdot)$ is learned incrementally, then each individual evaluation of the transformation will require either re-estimation of category means or mixture centers, or re-transformation of all exemplars in the dataset, both of which seem psychologically implausible and computationally intractable.
Alternatively, we can parameterize our likelihood function for each category $i$ by a set of $K$ learnable centers $z_i = (z_1,...,z_K)$ for category $i$, and take the likelihood function $p(\phi(x)|c_i, z_i))$ to be the Gaussian kernel density estimator proposed above, with kernels at locations $z_i$ for category $i$. 
The task of category learning can therefore be re-framed as inferring the probability 
\begin{equation}
    \argmax_{\phi, z}p(c_i\,|\,\phi(x))\propto p(\phi(x)|c_i, z_i)),
\end{equation}
When $K = 1$, the number of centers per category corresponds to learned prototypes for each category. Assuming an equal number of stimuli per class, when $K = \frac{N}{C}$ there are as many centers as stimuli per class, corresponding to learned exemplars. For $C$ total classes, when $\frac{N}{C} > K > C$, there are an arbitrary number of centers, and while they could be thought of as subsets of exemplars, it may be most appropriate to think of them as the centers of a Gaussian mixture density. In general, we are now interested in the probability
%
\begin{equation}
    p(c_i\,|\,\phi(x),z_i) \propto \sum_{z^\prime \in z_i} -d(\phi(x),\,z^\prime, {\bf I})^2.
\end{equation}
Note that exponentiation and normalization are omitted for clarity, since the proportional relationship still holds. When $K = 1$, the sum contains only a single term, corresponding to a Gaussian density (prototype model), when $\frac{N}{C} > K > C$, the sum corresponds to a Gaussian mixture density with uniform weights, and when $K = \frac{N}{C}$, the sum corresponds to a Gaussian-kernel density (exemplar model).

\subsection{Deep Categorization Models}

While transformation $\phi(\cdot)$ and centers $z_i$ could be estimated through a variety of methods, we choose to express both using a deep convolutional neural network \citep[CNN;][]{krizhevsky2012imagenet} with parameter matrices ($\theta_1$, $\theta_2$), which among other reasons are highly scalable and efficient for featurizing image inputs. The transformation $\phi_{\theta_1}\!$ is instantiated as a convolutional neural network and $Z_{\theta_2}\!$ parameterizes a final classification layer via learned centers $z_i$ for each category. A visualization of our deep categorization models is shown in Figure \ref{fig:diagram}. Particular variants of the model are explained in detail below.

\paragraph{Deep Prototype Models.} 
In the case of Gaussian likelihoods, the covariance matrix $\Sigma_j$ determines whether the distance function is Euclidean or Mahalanobis. This difference corresponds to constraints on the possible shape of the Gaussian category densities that will be estimated. When $\Sigma_j$ is the identity ${\bf I}$ (denoted as $\Sigma_{\bf I}$ hereafter), we assume that category distributions are isometric Gaussians (\textit{i.e.,} $d$-dimensional spheres in the stimulus or feature space). When the diagonal of $\Sigma_j$ is a scalar multiple of ${\bf I}$ (\textit{i.e.,} the diagonal is filled by the same constant that does not need to be $1$), denoted as $\Sigma_C$ hereafter, the size of the sphere and variance of the category is allowed to vary across categories. Finally, when the elements of the diagonal of $\Sigma_j$ can vary freely, denoted as $\Sigma_A$ hereafter, the category distributions are allowed to stretch into $d$-dimensional ellipsoids that are aligned with the axes of the stimulus space. An additional case is also possible, wherein $\Sigma_j$ is a full covariance matrix, and dimensions of the Gaussian are allowed to be correlated. However, we leave this much more expressive variant to future work.

Notably, DPMs are formally similar to a training objective called \textit{center loss} \citep{wen2016discriminative}, although the latter re-estimates class means for each mini-batches during training. They are also related to prototypical networks \citep{snell2017prototypical,scott2019stochastic}, which learn representations such that averages of exemplars from unseen classes support few-shot learning.

\paragraph{Deep Gaussian Mixture Models.}
DGMMs were implemented as multi-center versions of the DPMs described above, but distances are summed across all centers for each class, and the distance function is always Euclidean. This model contrasts with related ones that are instead trained asynchronously \citep{variani2015gaussian}, encourage large margins \citep{wan2018rethinking}, or focus on multiple layers of latent variables \citep{van2014factoring,viroli2019deep}.

\paragraph{Deep Exemplar Models.}
DEMs were implemented as DGMMs with as many centers as training datapoints. Exemplars were initialized as the initial random projections (a result of random starting weights) of the training data through the untrained neural network. It should be noted that another popular neural network model of category learning (which also does not incorporate feature learning) known as ALCOVE \citep{kruschke92} is a special case of our models where centers are arranged in a grid or randomly initialized in a fixed stimulus space, but not trained. DEMs are also related to deep Gaussian process hybrid deep networks \citep{bradshaw2017adversarial}, which are fitted with stochastic variational inference.


\input{tables}

\section{Methods}

\paragraph{Dataset.}
While there are a variety of large natural image datasets from computer vision that may be appropriate for training, \citep[\textit{e.g.,}][]{imagenet,barbu2019objectnet}, we focus on \texttt{CIFAR-10} \citep{krizhevsky2009cifar}, a popular benchmark; albeit one with relatively small images in terms of pixel size. We choose \texttt{CIFAR-10} both because it allows us to train many hundreds of models and replications to convergence in a reasonable amount of time (about 5-7 hours each), and because it allows for a rigorous comparison to an existing, large human behavioral dataset (see \textbf{Evaluation} below). \texttt{CIFAR-10} contains $5000$ training and $1000$ validation images for each of ten categories: \textit{airplane}, \textit{automobile}, \textit{bird}, \textit{cat}, \textit{deer}, \textit{dog}, \textit{frog}, \textit{horse}, \textit{ship}, and \textit{truck}.

\paragraph{CNN Architectures.}
To learn transformation $\phi_{\theta_1}\!$ over input images, we employ two architectures: ResNet \citep{he2016deep} and All-CNN \citep{springenberg2014striving}. All-CNN is a surprisingly parsimonious (\textit{i.e.,} uses only convolutional layers), low-parameter, stable, high-accuracy network that can be trained quickly. Our only architectural modification is the addition of batch normalization layers after each rectified linear (ReLU) activation function on top of each convolutional layer, which significantly reduced training time. Including the new layers before the ReLUs was less effective in our tests. For ResNet, we use the 20-layer, \textit{v2} version, which is by no means the best of its cousins, but has a high size-to-accuracy ratio. The final representation layers in All-CNN and ResNet are 192- and 256-dimensional respectively. We also use each unmodified architecture as a baseline model, with the original fully-connected classification layer at the end.

For both networks, we do not mirror the original training schemes exactly, which involve different optimizers and complicated learning rate schedules that would interact with our new architectural changes; instead, we hold them constant to reduce the search space. In particular we used a simple gradient descent optimizer with a learning rate of $0.01$, decay value of 1e-6, and Nesterov momentum with a value of $0.9$. These values were chosen before any baselines or novel models were trained.

\paragraph{Training.}
We train each of our models in the typical supervised manner. While categorization models in psychology are typically fit directly to human behavior, we follow the machine learning paradigm of presenting only single labels for each image. This mimics a type of supervision that humans receive, since humans do not learn categories by fitting to their own behavior. As explained below, we evaluate our models' fit to humans \textit{after} learning, requiring that our models learn to behave like humans without being explicitly trained to do so.

\paragraph{Evaluation.}
Following standard practice in machine learning, we evaluate the accuracy of our models on a held out set of images and ground-truth labels (in our case, the \texttt{CIFAR10} validation set described above). This is mostly to ensure that our baselines are valid and our models are competitive in this regard. However, we are mainly interested in fit to human behavior. \cite{Peterson_2019_ICCV} provide a useful dataset in this regard, which measures human uncertainty for each image of the $10,000$-image \texttt{CIFAR-10} validation, which they call \texttt{CIFAR-10H}.

\texttt{CIFAR-10H} contains 10-way classification judgments from $50$ subjects for each image (over $500,000$ total judgments), yielding probability distributions that reflect human uncertainty at the image level. For example, for a particular image, which are small and often somewhat ambiguous, 60\% of subjects may have chosen \textit{dog}, and the other 40\% may have chosen \textit{cat}. These proportions are directly comparable to the output of a network when expressed as probabilities. The authors validate this dataset by showing that training directly on it (as opposed to typical ``one-hot'' vectors) generalizes to human uncertainty for held out images, generalizes in terms of accuracy to a series of large and challenging held out validation datasets, and confers robustness to adversarial attacks. This both defines a crucial natural-image benchmark for predicting human behavior, and also introduces a new metric for objectively evaluating image classification models. To apply this dataset as evaluation, crossentropy (error) between the probability distributions output by a trained model $y_{i}$ and the human probabilities $h_{i}$ for each image in the validation set is computed: $-\sum_{i=1}^{C} h_{i} \log \left(y_{i}\right)$.
%

\section{Results}

\subsection{Performance}
For all results, we report both the best (Table 1) and the averages (Table 2) of $10$ training runs, which exhibit the same pattern of results. 
%
We first note that all but one of our eight models outperform the standard convolutional neural network baseline for accuracy on \texttt{CIFAR-10} ground-truth labels, with the Deep Gaussian Mixture Model scoring highest. As accuracy gains not trivial to obtain for well-studied architectures, these are encouraging results. 

However, as we are primarily interested in providing high-quality models of human categorization for more complex stimuli, the main criterion of model success is the fit to human-derived labels for these images from \texttt{CIFAR-10H}. Here we also find that all of our models outperform the neural network baselines, but with a much greater improvement, particularly for the DGMMs and DEMs. For comparison, in the study that introduced this \texttt{CIFAR10-H}, a ResNet architecture pretrained on \texttt{CIFAR10} and further fine-tuned on subsets of the \texttt{CIFAR10-H} human label distributions achieved a cross entropy loss of 0.35 \citep{Peterson_2019_ICCV}. Our models achieve nearly the same loss with none of the extra supervision, in most cases with lower accuracy (the authors do extensive hyperparameter search), and without any additional fine-tuning of hyper-parameters.

\begin{figure} 
  \centering
  \includegraphics[width=0.99\linewidth,trim={6mm 1mm 17mm 12mm},clip]{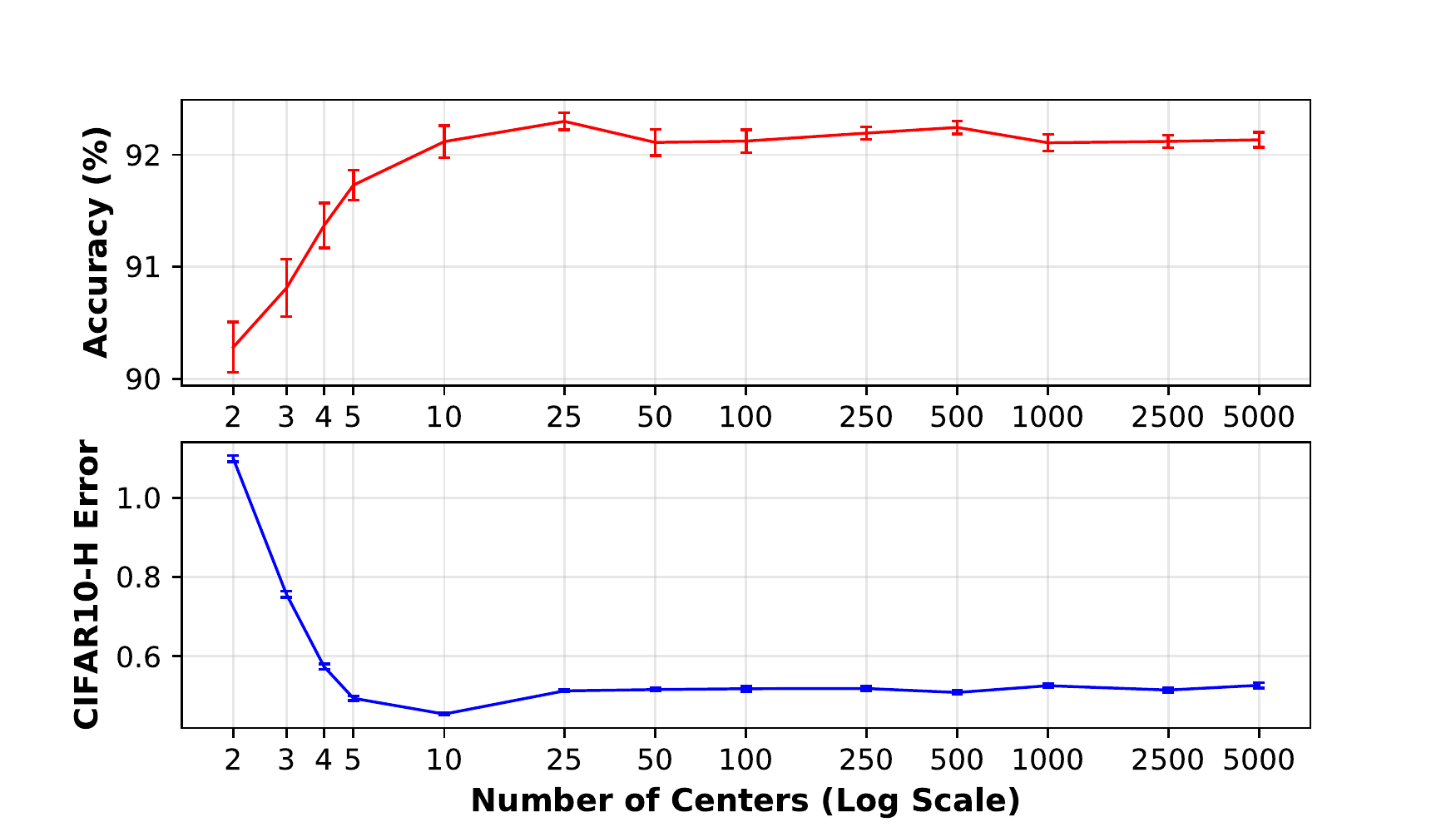}
  \vspace{-2mm}
  \caption{Accuracy and human fit as no. model centers varies.}
  \vspace{-1mm}
  \label{fig:n_centers}
\end{figure}

\subsection{Optimal Number of Centers}

In the previous section, we make a distinction between models with single clusters for each category (prototypes), models with as many clusters as training datapoints (exemplars), and anything in between (mixture centers). For DGMMs, we reported only the best models. To explore the optimal number of centers for predicting human uncertainty (and obtaining highest accuracy) more finely, we vary the number of centers from $2$ to $5000$ for All-CNN and plot them against each performance metric in Figure \ref{fig:n_centers}, averaged over $10$ runs. We first note that both accuracy and fit to humans are a smooth function of the number of centers and exhibit a clear maximum. To maximize accuracy, roughly $25$ centers are optimal; to maximize fit to human uncertainty, less than half of that are needed (roughly $10$). This suggests that both can be maximized given some number of centers $10 < M < 25$, and provides clearer evidence that GMM models of category learning, a middle ground between infamously opposed prototype and exemplar models, may best model joint human feature and category representations for natural images.

\subsection{Accuracy Versus Fit to Humans}
\begin{figure}
  \centering
  \includegraphics[width=0.91\linewidth,trim={2mm 7mm 15mm 16mm},clip]{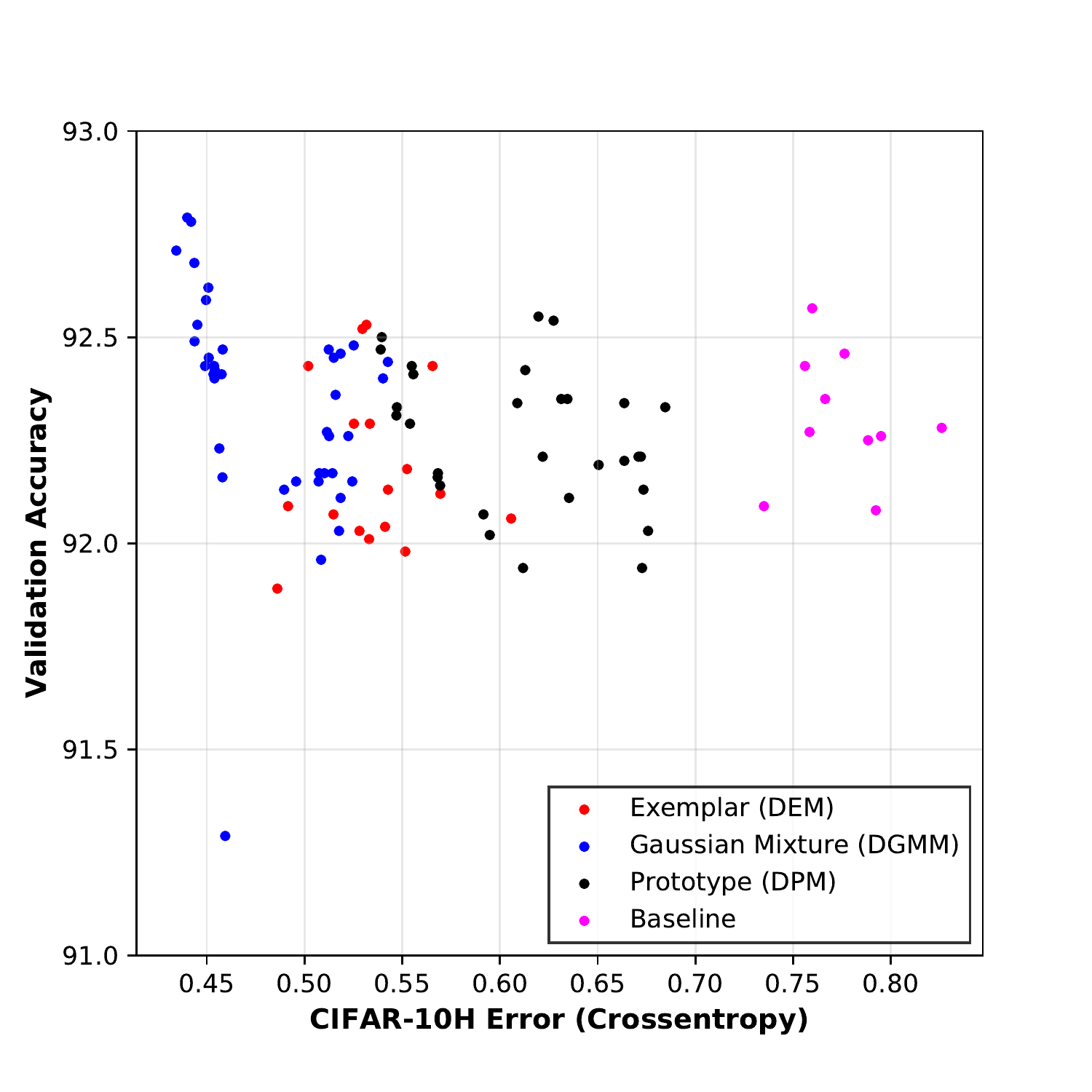}
  \vspace{-2mm}
  \caption{Validation accuracy $\times$ fit to human uncertainty.}
  \label{fig:acc_by_loss}
\end{figure}

Compared to baseline, our models are both more accurate and a much better fit to human uncertainty. While our models contain more parameters, the complexity (\textit{i.e.,} parameter count) of each model does not appear to be related to performance: the most complex DPMs (\textit{i.e.}, with axis-aligned covariance matrices) and the most complex multi-center models (\textit{i.e.,} DEMs) were not the best models. The optimal number of centers was also quite modest as we saw in the previous section. Further, accuracy does not appear to be related to human fit at all. In Figure \ref{fig:acc_by_loss}, we plot accuracy against error on \texttt{CIFAR10-H} for all $90$ All-CNN models that we trained and find essentially no discernible relationship. Instead, fit to humans appears to be solely a function of categorization model (identified by color).

\subsection{Visualizing Category Representations}

\begin{figure}
  \centering
  \includegraphics[width=1.0\linewidth,trim={0mm 0mm 0mm 0mm},clip]{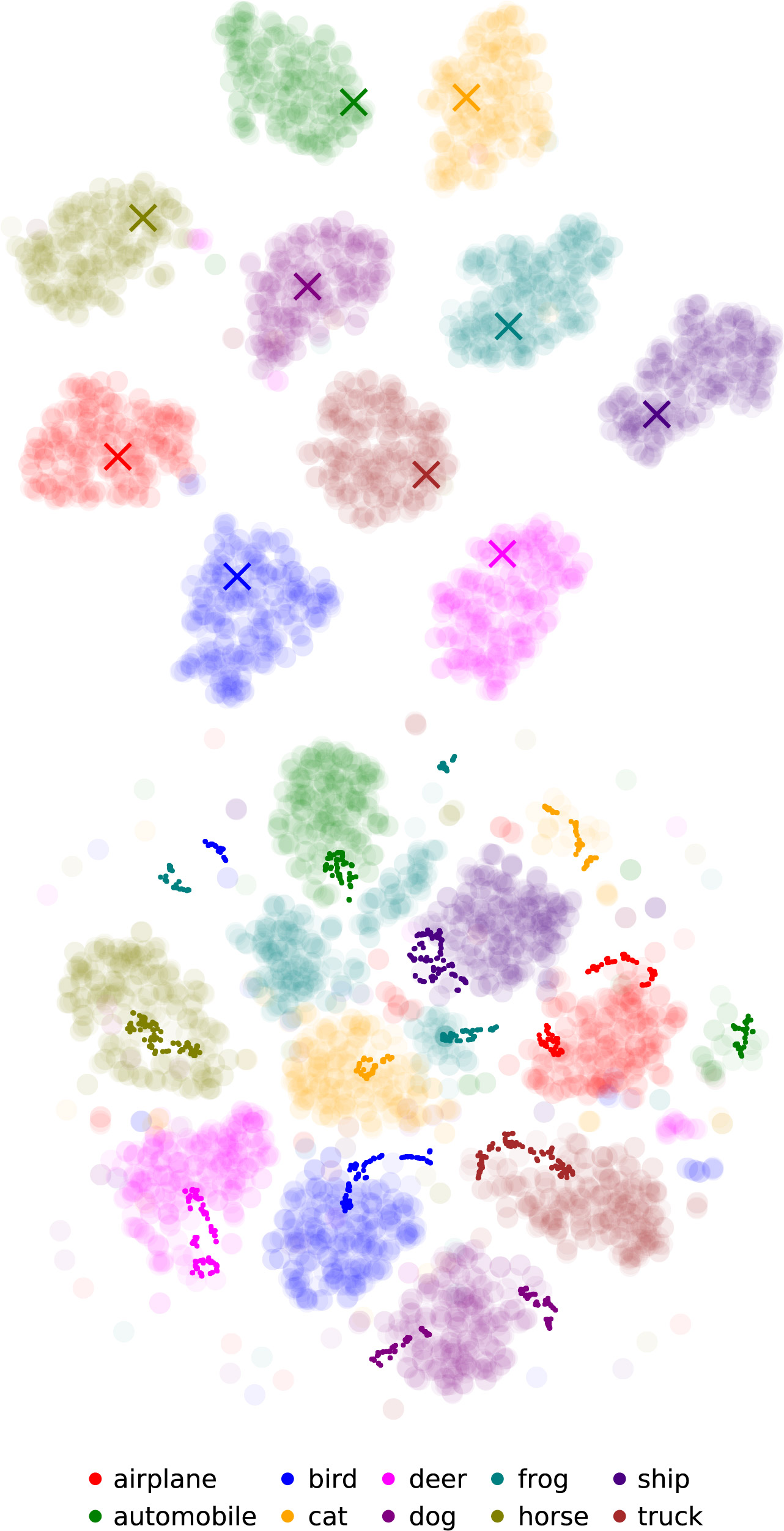}
  \caption{t-SNE embeddings of 3000 random stimuli and all per-category centers for a DPM (top) and DGMM (bottom).}
  \label{fig:tsne}
\end{figure}

Lastly, we visualize representations and centers using t-SNE for two illustrative model classes: identity-covariance DPMs, with a single center per class, and a DGMM with 100 centers per class. For the DPMs we expect to see a single point (dark color) representing the prototype for each category near the center of its members. For DGMMs, there are two possibilities: (1) a degenerate case where the model learned many near duplicates of the mean of the category, or (2) a more interesting case where centers have an arbitrary nontrivial distribution that supports a nonlinear classification boundary. The results are shown in Figure \ref{fig:tsne}. DPMs appear largely unimodal, as expected, and DGMMs indeed exhibit an interesting and non-degenerate pattern: centers are reasonably spread out, sometimes in nonlinear shapes or multimodal clusters, and not always located at the center of class members. The distribution of category members is also multimodal: subsets of centers are sometimes located distantly from the others in order to accommodate these flexible distributions.

\section{Discussion}
Traditional models of category learning in psychology are evaluated in the context of underlying stimulus representations that are chosen independently, which fails to model their interaction, and may significantly bias model comparison. We directly address this issue by learning the most appropriate representation for each categorization strategy.

Our approach appears to have been successful in that DCMs can be trained in a stable manner with no apparent drawbacks. Even networks that must employ up to $5000$ centers per category had no trouble learning useful stimulus representations in parallel. Moreover, nearly all of our models exhibit patterns of uncertainty in their post-training predictions that match people, and most of them improve accuracy, which is no easy feat given a fixed, well-known architecture in computer vision. Since we expected the former results but not the latter, we did not aim to augment the absolute state-of-the-art CNNs, and so cannot yet fully assess the utility of our models on that tier. However, the results we obtained were remarkably consistent across our ten training runs. The pattern of results reported in this paper were obtained in one large training batch, and were not revised to find hyperparameters more advantageous to our models' performance.

While the accuracy gains obtained by our models were modest, improvements in fit to human uncertainty were relatively large compared to previous work \citep{Peterson_2019_ICCV} that required extensive human-based supervision to obtain comparable results. This result is important because agreement between human and model uncertainty patterns is both a better predictor of robustness and generalization than accuracy \citep{Peterson_2019_ICCV}, and a crucial verification of psychological theories of category learning that posit explicit category representation. Although producing datasets like \texttt{CIFAR10-H} is not as straightforward for datasets of larger images with less ambiguity (and therefore less human uncertainty), the success of our category learning models on this dataset provides way to potentially improve network performance on any dataset, since fitting directly to human behavior is not necessary.

\section{Conclusion}
In the current work, we have proposed a set of models that generalize several seminal accounts of categorization to include feature learning, making use of deep neural networks as our workhorses. One advantage of doing so is that it enables us to study more complex and naturalistic stimulus sets, while ensuring stimulus and category representations cohere well with inaccessible human psychological representations. We see such integrations as the next step in the synergistic relationship between modern machine learning and cognitive science. In particular, we see this successful experiment as motivation for cognitive modelers to take inspiration from scalable methods, and as evidence to machine learning practitioners that insights into human cognition can tangibly inform the design of AI.

\section{Acknowledgements}
This work was supported by the National Science Foundation (grant number 1718550), as well as the Office of Undergraduate Research, the Center for Statistics and Machine Learning, the Program in Cognitive Science, and the  School of Engineering and Applied Sciences at Princeton University.

\nocite{shepard87}

\renewcommand{\bibliographytypesize}{\small}
\bibliographystyle{apacite}

\setlength{\bibleftmargin}{.125in}
\setlength{\bibindent}{-\bibleftmargin}

\bibliography{refs}

\end{document}

%% file: tables.tex
\begin{table*}[ht!]
\centering
\resizebox{\textwidth}{!}{%
\begin{tabular}{lcc||cccccc||cc||cc}
\hline
     & \multicolumn{2}{c}{Baseline} & \multicolumn{6}{c}{Prototype (DPM)}                                         & \multicolumn{2}{c}{DGMM} & \multicolumn{2}{c}{Exemplar (DEM)} \\ \hline
     & \textit{ResNet}       & \textit{AllCNN}      & \multicolumn{3}{c}{ResNet}           & \multicolumn{3}{c||}{\textit{AllCNN}}           & \textit{Resnet}               & \textit{AllCNN}               & \textit{ResNet}          & \textit{AllCNN}         \\
     &               &              & $\Sigma_{\textbf{I}}$ & $\Sigma_C$ & $\Sigma_A$ & $\Sigma_{\textbf{I}}$ & $\Sigma_C$ & $\Sigma_A$ &                      &                      &                  &                 \\
\texttt{CIFAR-10 } (Accuracy)  & 90.3          & 92.3         & 90.3       & 90.3       & 90.4       & 92.3       & 92.3       & 91.8       & 90.4 & \textbf{92.5}   &         90.2         &        92.1         \\
\texttt{CIFAR-10H} (Error) & .88           & .78          & .72        & .77        & .84        & .55        & .63        & .67        & .72                  & \textbf{.45}      &    .74              &        .57         \\ \hline
\end{tabular}
}
\label{table-means}
\vspace{-2mm}
\caption{Mean \texttt{CIFAR-10} validation accuracy and \texttt{CIFAR-10H} error (crossentropy; fit to human uncertainty) for each model.}
\end{table*}

\begin{table*}[ht!]
\centering
\resizebox{\textwidth}{!}{%
\begin{tabular}{lcc||cccccc||cc||cc}
\hline
     & \multicolumn{2}{c}{Baseline} & \multicolumn{6}{c}{Prototype (DPM)}                                         & \multicolumn{2}{c}{DGMM} & \multicolumn{2}{c}{Exemplar (DEM)} \\ \hline
     & \textit{ResNet}       & \textit{AllCNN}      & \multicolumn{3}{c}{ResNet}           & \multicolumn{3}{c||}{\textit{AllCNN}}           & \textit{Resnet}               & \textit{AllCNN}               & \textit{ResNet}          & \textit{AllCNN}         \\
     &               &              & $\Sigma_{\textbf{I}}$ & $\Sigma_C$ & $\Sigma_A$ & $\Sigma_{\textbf{I}}$ & $\Sigma_C$ & $\Sigma_A$ &                      &                      &                  &                 \\
\texttt{CIFAR-10 } (Accuracy)  & 90.3          & 92.3         & 90.4       & 90.5       & 90.5       & 92.4       & 92.4       & 91.9       & 90.8 & \textbf{92.8}   &         90.6         &        92.5         \\
\texttt{CIFAR-10H} (Error) & .88           & .76          & .72        & .77        & .84        & .53        & .63        & .67        & .71                  & \textbf{.43}      &    .72              &        .48         \\ \hline
\end{tabular}
}
\label{table-highs}
\vspace{-2mm}
\caption{Best \texttt{CIFAR-10} validation accuracy and \texttt{CIFAR-10H} error (crossentropy; fit to human uncertainty) for each model.}
\end{table*}